\documentclass[sigconf]{acmart}
\usepackage{multirow}
 \usepackage{subfig}

\usepackage{amsmath}
\usepackage[british]{babel}
\usepackage[autostyle]{csquotes}
\AtBeginDocument{%
  \providecommand\BibTeX{{%
    \normalfont B\kern-0.5em{\scshape i\kern-0.25em b}\kern-0.8em\TeX}}}

\setcopyright{acmcopyright}
\copyrightyear{2020}
\acmYear{2020}
\setcopyright{acmcopyright}\acmConference[CIKM '20]{Proceedings of the 29th ACM International Conference on Information and Knowledge Management}{October 19--23, 2020}{Virtual Event, Ireland}
\acmBooktitle{Proceedings of the 29th ACM International Conference on Information and Knowledge Management (CIKM '20), October 19--23, 2020, Virtual Event, Ireland}
\acmPrice{15.00}
\acmDOI{10.1145/3340531.3411994}
\acmISBN{978-1-4503-6859-9/20/10}
\begin{document}
\fancyhead{}
\title{RKT : Relation-Aware Self-Attention for Knowledge Tracing}

\author{Shalini Pandey}
\email{pande103@umn.edu}
\affiliation{%
  \institution{University of Minnesota}
  \city{Twin Cities}
  \state{Minnesota}
  \country{USA}
  \postcode{55455}
}

\author{Jaideep Srivastava}
\email{srivasta@umn.edu}
\affiliation{%
 \institution{University of Minnesota}
  \city{Twin Cities}
  \state{Minnesota}
  \country{USA}
  \postcode{55455}
  }

\begin{abstract}
The world has transitioned into a new phase of online learning in response to the recent Covid19 pandemic. Now more than ever, it has become paramount to push the limits of online learning in every manner to keep flourishing the education system. One crucial component of online learning is Knowledge Tracing (KT). The aim of KT is to model student's knowledge level based on their answers to a sequence of exercises referred as interactions. Students acquire their skills while solving exercises and each such interaction has a distinct impact on student ability to solve a future exercise. This \textit{impact} is characterized by 1) the relation between exercises involved in the interactions and 2) student forget behavior. Traditional studies on knowledge tracing do not explicitly model both the components jointly to estimate the impact of these interactions. \par
In this paper, we propose a novel Relation-aware self-attention model for Knowledge Tracing (RKT). We introduce a relation-aware self-attention layer that incorporates the contextual information. This contextual information integrates both the exercise relation information through their textual content as well as student performance data and the forget behavior information  through modeling an exponentially decaying  kernel function. 
Extensive experiments on three real-world datasets, among which two new collections are released to the public, show that our model outperforms
state-of-the-art knowledge tracing methods. Furthermore,
the interpretable attention weights help visualize the relation between interactions and temporal patterns in the human learning process.

\end{abstract}




\maketitle

   \begin{figure*}[!t]
    {
       \includegraphics[width=0.9\textwidth]{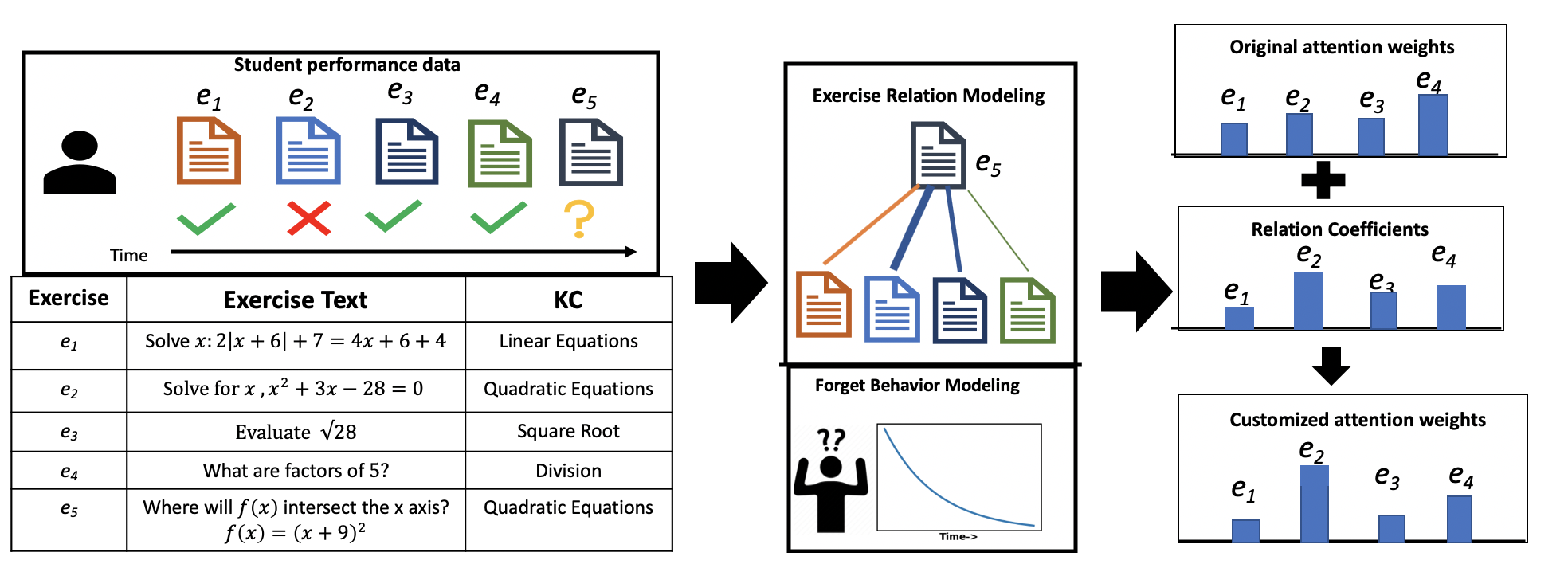}
     }
     \caption{Overview of RKT model: Leftmost figure shows a student performance data and table shows textual content and knowledge concepts of exercises which constitute the input of RKT. Middle figure shows the relation between exercises and forget behavior of student which serve as contextual information for RKT. Rightmost figure shows that contextual information encoded as relation coefficients informs the attention weight to revised attention weights. }
     \label{motivation}
   \end{figure*}
\section{Introduction}
Real-world education service systems, such as massive open online courses (MOOCs) and online platforms for intelligent tutoring systems on the web offers millions of online courses and exercises which have attracted attention from the public ~\cite{anderson2014engaging, lan2014time}. These online systems 
allow students to learn and do their exercises independently, and at their own pace ~\cite{corbett1994knowledge}. However, such a system requires a mechanism to help students realize their strengths and weaknesses so that they can practice accordingly. In addition to helping students, the mechanism can aid the teachers and system creators to proactively suggest remedial material and recommend exercises based on student needs ~\cite{kuh2011student}. 
For developing such a mechanism, knowledge tracing (KT) is considered to be crucial and is defined as the task of modeling students' \textit{knowledge state} over time ~\cite{corbett1994knowledge}.  It is an inherently difficult problem as it is dependent on the factors such as complexity of the human brain and ability to acquire knowledge ~\cite{piech2015deep}.\par

Figure ~\ref{motivation} shows an example of a student solving exercises sequentially. When the student encounters a new exercise (e.g. \textquote{$e_5$}) she applies her knowledge corresponding to the Knowledge Concept (e.g., \textit{Quadratic Equations}) to answer it. The mastery of a particular KC is determined by the past interactions which have a distinct impact on the target KC.  Besides, the impact is distinct under different circumstances. Typically, two factors account for determining the impact of past interactions in the prediction task: (1) exercise-relation (reflecting the relation between past exercises and the new exercise ), and (2) the time elapsed since the past interactions.  Intuitively, if the two exercises in the interactions are related to each other then the performance on one affects the other. Additionally, the knowledge gained while solving an exercise in the interaction decays with time, which is attributed to the forget behavior of students. It is important to use this information to contextualize the KT models. 
 
To model the evolution of student knowledge with interaction, Hidden Markov Models were traditionally used in Bayesian Knowledge Tracing (BKT) \cite{corbett1994knowledge} and its variants ~\cite{yudelson2013individualized, baker2009state}. Recently, the progress in sequential modeling using deep learning has inspired  Deep Knowledge Tracing (DKT) ~\cite{piech2015deep}, Dynamic Key-Value Memory Network (DKVMN) ~\cite{zhang2017dynamic} and Self-attentive Knowledge Tracing (SAKT) ~\cite{pandey2019self}  that are designed to capture long term dependencies between interactions.  Models such as ~\cite{kaser2014beyond, chen2018prerequisite} have shown the importance of explicitly incorporating the relations between KCs as input to the KT model. In particular,  ~\cite{kaser2014beyond} uses Dynamic Bayesian Network to model the pre-requisite relations between KCs while ~\cite{chen2018prerequisite} incorporate the same in DKT model.  However, they assume that the relation between KCs is known apriori. In fact, manual labeling of relations is labor-intensive work. To automatically estimate the relations between exercises,  ~\cite{lan2014sparse} estimates a mapping between each exercise and corresponding KCs and considers the exercise belonging to the same KC as related. While, ~\cite{su2018exercise, liu2019ekt} leverage the textual content of exercises to model semantic similarity relation between exercises. However, these models do not take into account temporal component which affects the importance of past interactions, owing to the dynamic behavior of the student learning process.  \par
 The temporal factors in knowledge tracing have been addressed in ~\cite{nagatani2019augmenting,pelanek2015modeling, inproceedings}. These methods mainly focus on the time elapsed since the last interaction with the same KC or previous interaction without modeling the relation between exercises involved in those interactions. 
However, as discussed, the previous interactions have a distinct impact on prediction task which is attributed to both exercise relation and temporal dynamics of the learning. \par

 In this paper, we propose a novel Relation-aware self-attention model for Knowledge Tracing (RKT)  that adapts the self-attention \\~\cite{vaswani2017attention} mechanism for KT task.  Specifically, we introduce a relation-aware self-attention layer that incorporates the contextual information and meanwhile, maintains the simplicity and flexibility of the self-attention mechanism. To this end, we employ a representation to capture the relation information, called \textit{relation coefficients}. In particular, the \textit{relation coefficients} are obtained from \textit{exercise relation modeling} and \textit{forget behavior modeling}. The former extracts relation between exercises from their textual content and student performance data. While the latter employs a kernel function with a decaying curve with respect to time to model student tendency to forget. Our experiments reveal that our model outperforms state-of-the-art algorithms on three real-world datasets. Additionally, we conduct a comprehensive ablation study of our model show the effect of key components and visualize the attention weights to qualitatively reveal the model’s behavior.\\
The contribution of our paper are:
\begin{itemize}
 \item We argue that each interaction in the sequence
has an adaptive impact on future interaction, where both
the relation between the exercises and the forgetting
behavior should be taken together into consideration.
\item We develop a method to learn the underlying relations between exercises using the textual content and student performance on the exercises which have not been explored before. 
\item We customize the self-attention model to incorporate the contextual information, thus enabling a fundamental adaptation of the model for KT. 
   \item  We perform extensive experiments on three real-world datasets and also illustrate that our model in addition to showing superior performance, provides an explanation for its prediction.
\end{itemize}
\section{Related Work}
\subsection{Cognitive Diagnosis}
Cognitive models refer to the models designed to discover latent mastery of each student on defined knowledge points.  Widely-used approaches
could be divided into two categories: one-dimensional models and multi-dimensional models. Among these models, Rasch model ~\cite{reise2014item} (also known as 1PL IRT) is a typical one-dimensional model and computes the probability of getting	an exercise correct using logistic regression based on student's ability and exercise (item) difficulty. To improve prediction results, other one-dimensional models include additive factor models~\cite{pavlik2009performance, cen2006learning}  which assumed KCs "additively" affect performance. These models include a student's proficiency parameter to account for the variability in student's learning abilities. Comparatively, multi-dimensional models, such as Deterministic Inputs, Noisy-And gate model, characterized students by a binary latent vector which described whether or not she mastered the KCs with the given Q-matrix prior ~\cite{de2011generalized}.\par
Similar to cognitive models, RKT also models the affect of past interactions on student performance. However, modeling human knowledge from past interaction is a complex task and we leverage the attention mechanism to capture the complexity involved in dynamics of past interactions for prediction task.
\subsection{Knowledge Tracing}
The KT task evaluates the knowledge state of a student based on her performance data. A Hidden Markov based model, BKT, was proposed in \cite{corbett1994knowledge}.  It models latent knowledge state of a learner as a set of binary variables, each of which represents understanding or non-understanding of a single concept. A Hidden Markov Model (HMM) is used to update the probabilities across each of these binary variables, as a student answers exercises. Further extension of BKT includes, incorporating individual student's prior knowledge~\cite{yudelson2013individualized}, slip and guess probabilities for each concept \cite{baker2009state} and the difficulty of each exercise~\cite{pardos2011kt}. Some approaches~\cite{thai2010recommender, toschercollaborative} use factorization methods to map each student into a latent vector that depicts her knowledge state. To capture the change of student's knowledge evolution over time, ~\cite{thai2011factorization} proposed a tensor factorization method by adding time as an additional dimension. 
Another line of research includes methods based on recurrent neural networks such as Deep Knowledge Tracing (DKT)~\cite{piech2015deep}, which exploits Long Short Term Memory (LSTM) to model student's knowledge state. Deep Knowledge Tracing plus (DKT+)~\cite{yeung2018addressing} is an extension of DKT to address the issue faced by DKT such as not being able to reconstruct the input and predicted KCs not being smooth across the time. Dynamic Key-Value Memory Networks (DKVMN)~\cite{zhang2017dynamic} introduced a Memory Augmented Neural Network ~\cite{santoro2016meta} to solve KT  with key being the exercises practices and values being the mastery of students.  Recently, Self-Attentive Knowledge Tracing (SAKT) ~\cite{pandey2019self} model was developed that first identifies the KCs from the student's past activities that are relevant to the target KC for which performance is to be predicted. SAKT then utilizes the information of performance on the past KCs to predict the student mastery at the next KC. \par 
Our method is an extension of SAKT such that we also take into account the relations between exercises involved in the interactions and time elapsed since the last interaction to inform the self-attention mechanism.
\subsection{ Relation Modeling in KT}
\textbf{Exercise Relation Modeling:} Exercise Relation Modeling has been widely studied in the educational psychology. Some researchers have utilized Q-matrix to map exercises with Knowledge Concepts ~\cite{barnes2005q, de2011generalized}. Two exercises are related if they belong to the same KC. 
In addition to Q-matrix based method, recently researchers have started to focus on deriving relations between exercises using the content of exercises. For example, ~\cite{liu2018finding, liu2019ekt,su2018exercise} utilize the content of exercises  to predict the relation between exercises. After predicting the semantic similarity scores between the exercises ~\cite{liu2019ekt, su2018exercise} use  these scores as attention weights to scale the importance of the past interactions. To the best of our knowledge, incorporating exercise relation modeling in KT is an under-explored area. To this end, we explored methods for modeling  exercise relations using textual content of exercises and student performance data. \par

\noindent \textbf{Forget Behavior Modeling:} There has been some research exploring the forget behavior of students ~\cite{ nagatani2019augmenting, chen2017tracking}. \textit{Forget curve theory} introduced in ~\cite{ebbinghaus2013memory} and employed in ~\cite{ chen2017tracking} which claims that student memory decays with time at an exponential rate and the rate of decay is determined by the strength of student cognitive abilities.
Recently,  DKT-Forget ~\cite{nagatani2019augmenting} introduce different
time-based features in DKT model. DKT-Forgetting considers repeated
and sequence time gap, as well as the number of past trials,
which is a state-of-the-art method with temporal information.
In our work, we take advantage of both exercise relation modeling and forget behavior modeling in KT task which has not been done before. 
\subsection{Attention Mechanism}
Attention mechanism~\cite{vaswani2017attention} is shown to be effective in tasks involving sequence modeling. The idea behind this mechanism is to focus on \textit{relevant} parts of the input when predicting the output. It makes the models often more interpretable as one can find the weights of specific input that resulted in making a specific prediction. It was introduced for machine translation task to retrieve the words in the input sequence for generating next word in the target sentence. Similarly, it is used in recommendation systems to predict the next item a person will buy based on his history of purchase. Some models ~\cite{ji2019sequential,yang2019context} have recognized that augmenting self-attention layer with contextual information improves the performance of the model. Such contextual information include the co-occurrence of items for item recommendation ~\cite{ji2019sequential} and syntactic and semantic information of a sentence for machine translation ~\cite{yang2019context}.
In our task, we use the self-attention mechanism to learn the attention weights corresponding to the previous interaction for predicting whether a student will provide correct answer to the next exercise. We then augment the exercise relations and forget behavior of students to enhance the model performance. \par

\begin{table}[]
\caption{Notations}
\label{notations}
\begin{tabular}{ll}
\toprule
Notations & Description\\
\hline

$E$                                                  & total number of exercises                                                    \\

$x_i $                    & $i$th interaction tuple of a student                                   \\
$d$                                                 & latent vector dimensionality                                                 \\
$e $                       & sequence of exercises solved by the student                                  \\

$\textbf{P} $                      & Positional embedding matrix                                                  \\
$\textbf{A}$ & exercise-exercise relation matrix\\
$\textbf{R}$ & relation coefficients of past interactions \\
$\hat{\textbf{x}}_i$ &$i$th interaction embedding \\
$\textbf{P}$ & Positional embedding matrix\\
$l$ & maximum sequence length\\
$\textbf{E}$ & Exercise embedding 
\\

$\textbf{X}$                        & Interaction sequence of a student: \\
&$(x_1, x_2, \ldots, x_i)$ \\
\bottomrule
\end{tabular}
\end{table}

   \begin{figure*}[!t]
    {
       \includegraphics[width=1\textwidth]{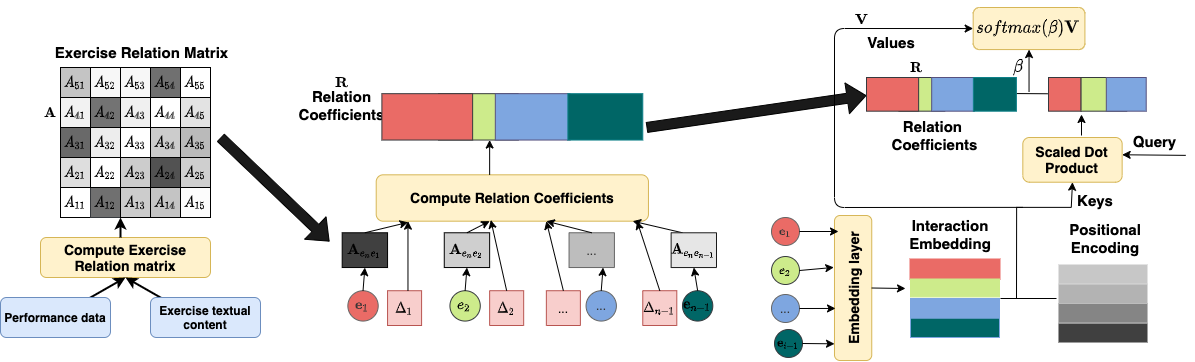}
     }
     \caption{The overall architecture of RKT. We first compute the exercise relation matrix $\textbf{A}$. Then we use $\textbf{A}$ to compute the relation coefficients based on the relation between past exercises $(e_1, e_2, \ldots e_{n-1})$ and the next exercise $e_n$ and the time elapsed since the interaction $(\Delta_1, \Delta_2, \ldots, \Delta_{n-1})$. The relation coefficients are propagated to the transformer model which modifies the attention weights to take into account the contextual information.  }
     \label{architecture}
   \end{figure*}
\section{Proposed Method}
 Knowledge Tracing  predicts whether a student will be able to answer the next exercise $e_{n}$ based on his/her previous interaction sequences ${X}=\{{x}_1, {x}_2, \ldots,{x}_{n-1}\}$. Each interaction is characterized by tuple $x_i = (e_i, r_i, t_i)$, where $e_i \in \{1, \ldots, E\}$ is the exercise attempted, $r_i \in \{0,1\}$ is the correctness of the student answer, and $t_i \in \mathbb{R}^+$ is the time at which the interaction occurred.  For accurate prediction, it is important to identify the underlying relation between $e_{n}$ attempted at time $t_n$ and the previous interactions. As shown in Figure ~\ref{motivation} the importance of a past interaction in predicting whether the student will be able to answer the next exercise correctly is determined by two factors: 1) the relation between the exercises solved in the past interaction and the next exercise, and 2) time elapsed since the past interaction. Motivated by this, we develop a Relation-aware Knowledge Tracing model which incorporates the relations as contextual information and propagates it to the attention weights computed using self-attention mechanism \cite{vaswani2017attention}. The updated attention weights are then used to compute the weighted sum of the representation of the past interactions which represents the output corresponding to the $n$th interaction.  To learn the parameters, we employ a binary cross entropy loss as our objective function.
 The mathematical notations used in this paper are summarized in Table ~\ref{notations}.
 
%

\subsection{Exercise Representation}

We learn a semantic representation of each exercise from its textual content. For this, we exploit word embedding technique and learn a function $f : M \rightarrow \mathbb{R}^d$ , where $M$ represents the dictionary of words and $f$ is a parameterized function which maps
words to $d$-dimensional distributed vectors. In the look-up
layer, exercise content are represented as a matrix of word embeddings. Then the embedding of an exercise $i$, $\textbf{E}_i \in \mathbb{R}^d$ is obtained by taking weighted combination of embedding of all the words present in the text of the exercise $i$ using Smooth Inverse Frequency (SIF) ~\cite{arora2016simple}. SIF downgrades unimportant words such as but, just, etc., and keeps the information that contributes the most to the semantics of the exercise. Thus, the exercise embedding for an exercise $i$ is obtained as:
\begin{equation}
    \textbf{E}_i = \frac{1}{|s_i|} \sum_{w\in s_i} \frac{a}{a+p(w)} f(w),
\end{equation}
\noindent where $a$ is a trainable parameter, $s_i$ represents the text of $i$th exercise, and $p(w)$ is the probability of word $w$. 
\subsection{Exercise-Relation Matrix Computation}
An important innovation of our model is that we explore methods of identifying the underlying relations between exercises. Since the relations between exercises are not explicitly known, we first infer these relations from the data and build a \textit{exercise relation matrix}, $\textbf{A}\in \mathbb{R}^{E\times E}$ such that $\textbf{A}_{i,j}$ represents the importance that performance on exercise $j$ has on the performance on exercise $i$.  We leverage two sources of information for discovering  the relations between exercises: student's performance data and textual content of exercises.  The former is used to capture the relevance of knowledge gained in solving exercise $j$ for solving exercise $i$, while the latter captures the semantic similarity between the two exercises. \par
We will now describe how learner's performance data can be used to obtain the  relevance of the knowledge gained from exercise $j$ to solve exercise $i$.
\begin{table}[]
\caption{A contingency table for two exercises $i$ and $j$.  }
\label{contingency}
\begin{tabular}{cccc|c}

\toprule
                            &           & \multicolumn{2}{c|}{exercise $i$} \\
                            
                            &           & incorrect      & correct   &total    \\
                            \midrule
\multirow{2}{*}{exercise $j$} & incorrect & $n_{00}$      & $n_{01}$ & $n_{0*}$    \\
                            & correct   & $n_{10}$      & $n_{11}$ & $n_{1*}$\\
                            \midrule
& total  & $n_{*0}$& $n_{*1}$& $n$\\
\bottomrule
\end{tabular}
\end{table}
We first build a contingency table as shown in table ~\ref{contingency} by considering only the pairs of $i$ and $j$, where $j$ occurs before $i$ in the learning sequence. If there are multiple occurrences of $j$ in the learning sequence before $i$, we only consider the latest occurrence.  Then, we compute the Phi coefficient which is popularly used as a measure of association for two binary variables. Mathematically the Phi coefficient that describes the relation from $j$ to $i$ is calculated as,
\begin{equation}
    \phi_{i,j} =\frac{n_{11}n_{00}-{n_{01}n_{10}}}{\sqrt{n_{1*}n_{0*}n_{*1}n_{*0}}} . 
\end{equation}
The value of $\phi_{i,j}$ lies between $-1$ and $1$ and a high $\phi_{i,j}$ score means students' performance at $j$ play an important role in deciding their performance at $i$. We choose Phi coefficients among other correlation metrics to compute the relation between exercises because: 1) it is easy to interpret, and 2) it explicitly penalizes when the two variables are not equal.
\par
Another source of data we use for computing relation between two exercises is the textual content of exercises which informs the semantic similarity of two exercises. We first obtain the exercise embedding of  $i$, $\textbf{E}_i$ 
and $j$, $\textbf{E}_j$ from section 3.1, then compute the similarity between exercises using cosine similarity of the embeddings.
Formally, similarity between exercises is calculated as:
\begin{equation}
    \text{sim}_{i,j}=\frac{\textbf{E}_i \textbf{E}_j}{||\textbf{E}_i||_2 ||\textbf{E}_j||_2}
\end{equation}

\par

Finally,  the relation of exercise $j$ with exercise $i$ is calculated as :
\begin{equation}
 \textbf{A}_{i,j}= 
\begin{cases}
    \phi_{i,j}+\text{sim}_{i,j} ,& \text{if } \text{sim}_{i,j}+\phi_{i,j}>\theta\\
    0,              & \text{otherwise},
\end{cases}
\end{equation}
where $\theta$ is a threshold that controls sparsity of relation matrix.  \par
\subsection{Personalized Relation Modeling}
Here we model the contextual information to compute the relevance of past interaction, represented as relation coefficients,  for predicting student performance at next exercise. Specifically, we incorporate the exercise relation  modeling and forget behavior modeling described below at this step.  \\
\textbf{Exercise Relation Modeling:} This component involves modeling the relation between exercises involved in interaction. Given the past exercises solved by a student, $(e_1, e_2, \ldots, e_{n-1})$ and the next exercise $e_n$ for which we want to predict its performance, we compute the  exercise-based relation coefficients   from the $e_n$th row of exercise relation matrix, $\textbf{A}_{e_n}$ as $\textbf{R}^E = [\textbf{A}_{e_n,e_1},\textbf{A}_{e_n,e_2},  \ldots,\textbf{A}_{e_n,e_{n-1}}]$. 

\noindent \textbf{Forget behavior modeling:} Learning theory has revealed that students forget the knowledge learnt with time ~\cite{averell2011form, ebbinghaus2013memory}, known as \textit{forgetting curve theory},  which plays an important role in knowledge tracing. Naturally, if a student forgets the knowledge gained after a particular interaction $i$, the relevance of that interaction for predicting student performance at the next interaction should be diminished, irrespective of the relation between exercises involved. The challenge is to identify the interactions whose knowledge the student has forgotten. Since students forget  with time,  we employ a kernel function that models the importance of interaction with respect to time interval. The kernel function is designed as an exponentially decaying curve with time to reduce the importance of interaction as time interval increases following the idea from forgetting curve theory. Specifically, given the time sequence of interaction of a student $\textbf{t} = (t_1,t_2, \ldots ,t_{n-1}) $ and the time at which the student attempts next exercise $t_{n}$, we compute the relative time interval between the next interaction and the $i$th interaction as $\Delta_{i}=t_{n}-t_i$. Thus, we compute forget behavior based relation coefficients, $\textbf{R}^T = [\exp(-\Delta_{1}/S_u), \exp(-\Delta_{2}/S_u), \ldots, \exp(-\Delta_{n-1}/S_u)]$, where $S_u$  refers to relative strength of memory of student $u$ and is a trainable parameter in our model. \par
Following ~\cite{yang2019context}, we also obtain revised importance of the past interaction by simply adding the weights obtained from individual sources of information. Thus, we compute the relation coefficients as
\begin{equation}
    \textbf{R} = \text{softmax}(\textbf{R}^E + \textbf{R}^T),
\end{equation}

The relation coefficient corresponding to more relevant interaction is higher.
\subsection{Input Embedding Layer}

The raw data of interactions only consists of tuple representing exercise, correctness and time of interaction. We need to embed this information of interactions and positions of interactions. To obtain an embedding of a past interaction $j$, $(e_j, r_j, t_j)$, we first obtain the corresponding exercise representation using Equation (1). To incorporate the correctness score $r_{j}$, we extend it to a feature vector $\textbf{r}_j = [r_j,  r_j, \ldots, r_j]  \in  \mathbb{R}^d$ and concatenate it to the exercise embedding. 
 Also, we define a positional embedding matrix as $\textbf{P} \in \mathbb{R}^{l\times 2d}$, to introduce the sequential ordering information of the interactions, where $l$ is the maximum allowed sequence length. The position embedding is particularly important in knowledge tracing problem because a student's knowledge state at a particular time instance should not show wavy transitions~\cite{yeung2018addressing}.\par

Afterward, we feed the inputs to RKT, and these inputs should convey the representation of interactions and positions
in the sequences.  Thus, the interaction embedding is obtained as:
\begin{equation}
    \hat{\textbf{x}}_j = [\textbf{E}_{e_j} \oplus \textbf{r}_j] + \textbf{P}_j
\end{equation}

Finally, the input interaction sequence is
expressed as $\hat{\textbf{X}}  = [\hat{\textbf{x}}_1, \hat{\textbf{x}}_2, \ldots \hat{\textbf{x}}_n]$ by combining the interaction embedding $\textbf{E}$, and the positional embedding $\textbf{P}$. 
\subsection{ Relation-Aware Self-attention Layer}
The core component of RKT is the attention structure
that incorporates relation structure.  For this, we modify the alignment score of the attention mechanism to attend more to the relevant  interactions identified by the relation coefficient, $\textbf{R}$. Let $\alpha$ be the attention weights learned using scaled dot-product attention mechanism ~\cite{vaswani2017attention} such that

\begin{equation}
    \alpha_{j} =  \frac{\exp(e_{j})}{\sum_{k=1}^{n-1} \exp(e_{k})}, e_{j}=\frac{\textbf{E}_{e_n}\textbf{W}^Q(\hat{\textbf{x}}_j\textbf{W}^K)^T}{\sqrt{d}},
\end{equation}
where $\textbf{W}^Q \in \mathbb{R}^{d\times  d}$ and $\textbf{W}^K 
\in \mathbb{R}^{d\times d} $ are  projection matrices for query
and key, respectively. Finally we combine the attention weights with the relation coefficients, by adding the two weights: 
\begin{equation}
    \beta_{j}=\lambda \alpha_{j}+(1-\lambda)\textbf{R}_{j},
\end{equation}
where $\textbf{R}_j$ is the $j$th element of the relation coefficient $\textbf{R}$.
\noindent We used addition operation to avoid any significant increase in computation cost. $\lambda$ is a tunable parameter. The representation of output at the $i$th interaction, $\textbf{o}\in \mathbb{R}^d$, is obtained by the weighted sum of linearly
transformed interaction embedding and position embedding:
\begin{equation}
    \textbf{o} = \sum_{j=1} ^ {n-1} \beta_{j}\hat{\textbf{x}}_j\textbf{W}^V,
\end{equation}
\noindent where $\textbf{W}^V\in \mathbb{R}^{d\times d}$ is the projection matrix for value space.\par
\textbf{Point-Wise Feed-Forward Layer:} We apply the PointWise Feed-Forward Layer (FFN) to the output of RKT by each position.  The FFN helps incorporate non-linearity in the model and considers the interactions between different latent dimensions. It consists of two linear transformations with a ReLU nonlinear activation function between the linear transformations. The final output of FFN is $\textbf{F} =  \text{ReLU}(\textbf{o}\textbf{W}^{(1)} + \textbf{b}^{(1)}) \textbf{W}^{(2)}+\textbf{b}^{(2)}$, where $\textbf{W}^{(1)} \in \mathbb{R}^{d\times d}$, $\textbf{W}^{(2)} \in \mathbb{R}^{d\times d}$ are weight matrices and $\textbf{b}^{(1)} \in \mathbb{R}^{d}$ and $\textbf{b}^{(2)} \in \mathbb{R}^{d\times d}$ are the bias vectors. \par
Besides of the above modeling structure, we added residual connections ~\cite{he2016deep} after both self-attention layer and Feed forward layer to train a deeper network structure. We also applied the layer
normalization ~\cite{ba2016layer} and the dropout
~\cite{srivastava2014dropout} to the output of each layer, following ~\cite{vaswani2017attention}.


\subsection{Prediction Layer}
Finally, to obtain student ability to answer exercise $e_n$  correctly, we pass the learned representation $\textbf{F} $ obtained above  through the fully connected network with Sigmoid activation to predict the performance of the student. 
\begin{equation}
    p = \sigma(\textbf{F}\textbf{W}+\textbf{b}),
\end{equation}
where $p$ is a scalar and represents the probability of student providing correct response to exercise $e_n$, and $\sigma(z) = 1/(1+e^{-z})$.

\subsection{Network Training} 
 Since the self-attention model works with sequence of fixed length, we convert the input sequence, $X = (x_1, x_2, \ldots, x_{|X|})$, into sequence of fixed length $l$ before feeding it to RKT. If the sequence length, $|X|$ is less than $l$,  we repetitively add a \textit{padding} to the left of the sequence. However, if $|X|$ is greater than $l$, we partition the sequence into subsequences of length $l$.
 The objective of training is to minimize the negative log likelihood of the observed sequence of student responses under the model. The parameters are learned by minimizing the cross entropy loss between $p$ and $r$ at every interaction.
\begin{equation}
    \mathcal{L} = -\sum_{i \in I} (r_i \log(p_i)+ (1-r_i)\log(1-p_i)),
\end{equation}
\noindent where $I$ denotes all the interactions in the training set.\footnote{The corresponding code and dataset available at https://github.com/shalini1194/RKT} 

\section{Experimental Settings}
In this section, we present our experimental settings to answer the following questions: \\
\textbf{RQ1} Can RKT outperform the state-of-the-art methods for Knowledge Tracing?  \\
\textbf{RQ2}: What is the influence of various components in the
RKT architecture?\\
\textbf{RQ3} Are the attention weights able to learn meaningful
patterns in computing the embeddings?\\
\begin{table}[]
\caption{Dataset Details}
\begin{tabular}{crrr}
\toprule
\multicolumn{1}{l}{}        & \multicolumn{1}{c}{ASSIST2012} & \multicolumn{1}{c}{Junyi} & \multicolumn{1}{c}{POJ} \\
\midrule
\# students                 & 39,364                         & 238,120                   & 22,916                  \\

\# exercises                & 58,761                         & 684                       & 2,751                   \\
\# Interactions             & 4,193,631                      & 26,666,117                & 996,240                 \\
Avg exercise record/student & 107                            & 111.99                    & 43.47                   \\
Duration of data collection & 365 days                       & 1095 days                 & 258 days     \\
\bottomrule
\end{tabular}
\label{dataset}
\vspace{-1.5em}
\end{table}
\subsection{Datasets}
To evaluate our model, we used three real-world datasets.
\begin{itemize}
\item \textbf{ASSISTment2012(ASSIST2012)\footnote{https://sites.google.com/site/assistmentsdata/home/2012-13-school-data-with-affect}:} This dataset is provided by ASSISTment online tutoring platform and is widely used for KT tasks. We also utilized the problem bodies to conduct our experiments. 
\item \textbf{JunyiAcademy (Junyi)\footnote{https://pslcdatashop.web.cmu.edu/DatasetInfo?datasetId=1275}} This dataset was collected by JunyiAcademy\footnote{https://www.junyiacademy.org/} in 2015 ~\cite{chang2015modeling}. The available dataset only contains the exercising records of students. To obtain the textual content we scraped the data from their website. Overall, this dataset contains 838 distinct exercises and we removed exercises which do not contain textual content. 
\item \textbf{Peking Online Judge (POJ)} This dataset is collected from Peking online platform of coding practices and consists of  computer programming questions.We scraped the publicly available data from the website~\footnote{http://poj.org/}. 
\end{itemize}
For all these datasets, we first removed the students who attempted fewer than two exercises and then removed those exercises which were attempted by  fewer than two students. 
The complete statistical information for all the datasets can be found in Table ~\ref{dataset}.
The code and dataset is available at  \textit{https://github.com/shalini1194/RKT}.

\subsection{Implementation Details}
\subsubsection{Word Embeddings} The first step in our method is to embed exercise content and initializing each word of the exercise content. All exercises are truncated to no more than $200$ words. However, mathematical exercises consists of words not found in traditional English articles such as, news. For example it is common to find formulas like "$\sqrt(x)+1$" in mathematical exercise which carry important information about the exercise. Therefore, to preserve the mathematics semantics, we transform each formula into its TEX code features ("$\sqrt(x)+1$" is transformed to " sqrt x + 1"). After initialization, each exercise is represented with sequence with vocabulary words and TEX tokens. The model is trained by embedding each word into an embedding vector with $50$ dimensions (i.e., d = $50$) by using word2vec ~\footnote{https://radimrehurek.com/gensim/models/word2vec.html}.\\
\subsubsection{Framework Setting} We now specify the network initializations
in our model. We set the model dimension  in self-attention as $64$ and the maximum allowed sequence length $l$ as $50$. The model is trained with a mini-batch
size of $128$. We use Adam optimizer with a learning rate of $0.001$.
The dropout rate is set to $0.1$ to reduce overfitting. The L2 weight
decay is set to $0.00001$. All the model parameters are normally initialized
with $0$ mean and $0.01$ standard deviation. The value of sparcity controlling threshold, $\theta$ used in Eq. (4) is 0.8 in our experiments. We trained the model with $80\%$ of the dataset and test it on the remaining. We perform 5-fold cross validation to evaluate all the models, in which folds are split based on
students. \par
\subsection{Metrics} 
The prediction of student performance is considered in a binary classification setting i.e., answering an exercise correctly or not. Hence, we compare the performance using the Area Under Curve (AUC) and Accuracy (ACC)  metric. 
Similar to evaluation procedure employed in ~\cite{nagatani2019augmenting, piech2015deep}, we train the model with the interactions in the training phase and during the testing phase, we update the model after each exercise response is received. The updated model is then used to perform the prediction on the next exercise.  Generally, the value $0.5$ of
AUC or ACC represents the performance prediction result
by randomly guessing, and the larger, the better.
\vspace{-2mm}
\subsection{Approaches}
\subsubsection{Knowledge Tracing (KT)}
We compare our model against the  state-of-the-art KT methods. 
\begin{itemize}
    \item \textbf{DKT} ~\cite{piech2015deep} : This is a seminal method that uses single layer LSTM model to predict the student's performance. In our implementation of DKT, we used norm-clipping and early stopping to improve the performance as has been employed in ~\cite{zhang2017dynamic}. 
    \item \textbf{SAKT} ~\cite{pandey2019self}This model employs self-attention mechanism ~\cite{vaswani2017attention} to assigns weights to the previously answered exercises for predicting the performance of the student on a particular exercise.  
    \item \textbf{DKVMN} ~\cite{zhang2017dynamic}: This is a Memory Augmented Recurrent Neural Network based method where in the relation between different KCs are represented by the \textit{key} matrix and the student's mastery of each KC by the \textit{value} matrix.  
    \item \textbf{DKT+Forget} ~\cite{nagatani2019augmenting} : This is an extension of DKT method which predicts student performance using both the student's learning sequence and fogetting behavior. 

    \item \textbf{EERNN} ~\cite{su2018exercise}: This model utilizes  both the textual content of exercises and student's exercising records to predict student performance. They use RNN as the underlying model to learn the exercise embedding and the student knowledge representation. Furthermore, they attend over the past interactions using the cosine similarity between the past interactions and the next exercise.
   
    \item \textbf{EKT} ~\cite{liu2019ekt}: This model is an extension of the EERNN model which also tracks student knowledge acquisition on multiple skills. Specifically, it models the relation between the underlying Knowledge Concepts to enhance the EERNN model.
 \end{itemize}

\section{Results and Discussion}
\subsection{Student Performance Prediction (RQ1)}
\begin{table}[]
\caption{Performance comparison. The best performing method  is boldfaced, and the second best method
in each row is underlined. Gains are shown in the last row.}
\begin{tabular}{c|r|r|r|r|r|r}
\toprule
\multicolumn{1}{l|}{} & \multicolumn{2}{|c|}{ASSIST2012}                    & \multicolumn{2}{|c|}{POJ}                           & \multicolumn{2}{|c}{Junyi}      
\\
\midrule
\multicolumn{1}{l|}{} & \multicolumn{1}{|l|}{AUC} & \multicolumn{1}{|l|}{ACC} & \multicolumn{1}{|l|}{AUC} & \multicolumn{1}{|l|}{ACC} & \multicolumn{1}{|l|}{AUC} & \multicolumn{1}{|l}{ACC} \\
\midrule
DKT                  & 0.712                   & 0.679                   & 0.656                   & 0.691                   & 0.814                   & 0.744                   \\
SAKT                 & 0.735                   & 0.692                   & 0.696                   & 0.705                   & 0.834                   & 0.757                   \\
DKVMN                & 0.701                   & 0.686                   & 0.704                   & 0.700                   & 0.822                   & 0.751                   \\
DKT+Forget           & 0.722                   & 0.685                   & 0.662                   & 0.700                   & 0.840                   & 0.759                   \\
EERNN                & 0.748                   & 0.698                   & 0.733                   & 0.720                   & 0.837                   & 0.758                   \\
EKT                  & \underline{0.754}                   & \underline{0.702}                   & \underline{0.737}                   & \underline{0.729}                   & \underline{0.842}                   & \underline{0.759}                  \\
RKT                  & \textbf{0.793}                   & \textbf{0.719}                   & \textbf{0.827}                   & \textbf{0.774}                   & \textbf{0.860}                   & \textbf{0.770}                   \\
\midrule
Gain\%               & 5.172                   & 2.422                   & 12.212                  & 6.173                   & 1.775                   & 1.050    \\
\bottomrule
\end{tabular}
\label{comparison}
\end{table}
Table ~\ref{comparison} shows the performance of all baseline methods and our RKT model. We have the following observations:\\
  Different kinds of baselines demonstrate noticeable performance gaps. SAKT model shows improvement over DKT and DKVMN model which can be traced to the fact that SAKT identifies the relevance between past interactions and next exercise. DKT-Forget further gains improvements most of the time, which demonstrates the importance of taking temporal factors into consideration. 
 Further, EERNN and EKT incorporate textual content of exercises to   identify which interaction history is more relevant and hence perform better than the those models which do not take into account these relations. RKT performs consistently better than all the
baselines. Compared with other baselines, RKT is able to explicitly captures the relations between exercises based on student performance data and text content. Additionally, it models learner forget behavior using a kernel function which is more interpretable and proven way to model human memory ~\cite{ebbinghaus2013memory} compared to DKT+forget model.  \par

Second, the performance gain is lowest for Junyi dataset. We believe that a possible reason of low improvement on Junyi is that since the number of exercises in Junyi is fairly small the relation between exercises can be modeled by sequential models such as RNN and self-attention mechanism. It does not need explicit relation learning based on the content.  \par 
We would also like to point out that, combining the model with contextual information in RKT does not lead to any significant increase in runtime of the model and it remains as scalable as SAKT model. SAKT and RKT are more scalable than other sequential models because of its parallelization capability ~\cite{pandey2019self}.

   \begin{figure*}[!t]
    {
       \includegraphics[width= 0.9\textwidth]{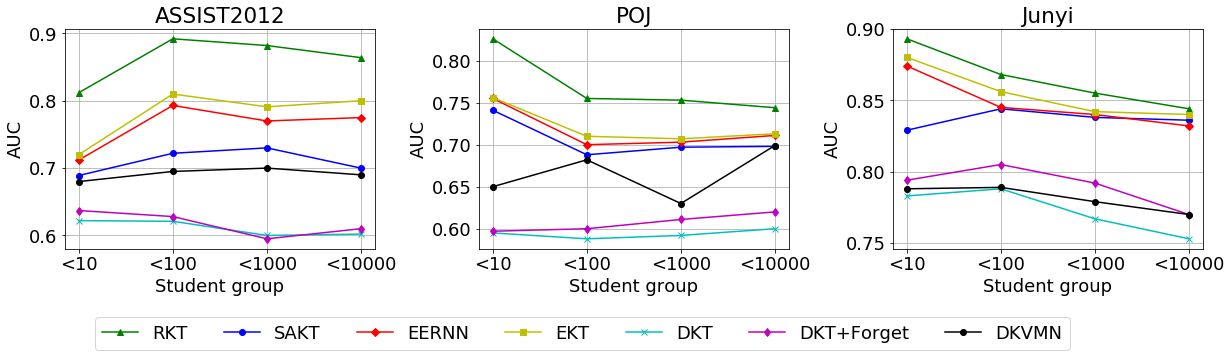}
}
\caption{Plot of prediction performance over different student groups based on sparsity of interaction levels. Our model, RKT  significantly outperforms every baseline.}
 \label{user_comp}

 \end{figure*}
\subsubsection{Performance comparison w.r.t. interaction sparsity}
One benefit of exploiting the relations between interactions is that it makes our model robust towards sparsity of dataset. Exploiting the relation between different exercises can help in estimating student performance at related exercises, thus alleviating the sparsity issue.

To verify this, we perform an experiment over student groups with
different number of interactions. In particular, we generate four groups  of students based on interaction number per user, thus generating groups  with less than $10$, $100$, $1000$, $10000$ interactions, respectively.
The performance of all the methods is displayed in Figure ~\ref{user_comp}.  
We find that RKT outperforms the baseline models in all the cases, signifying the importance of leveraging relation information for predicting performance. Also,  the performance gain of RKT for student groups with less number of interactions is more significant.
Thus, we can reach to a conclusion that RKT
which exploits the relation between interactions is effective for learning knowledge representation of students even with less interactions.  

 \subsection{Ablation Study (RQ2)}
 \begin{table}[]
\caption{Ablation Study}
\begin{tabular}{c|r|r|r|r|r|r}
\toprule
\multicolumn{1}{l}{} & \multicolumn{2}{|c|}{ASSIST2012}                    & \multicolumn{2}{|c|}{POJ}                           & \multicolumn{2}{|c}{Junyi}                         \\
\midrule
\multicolumn{1}{l}{} & \multicolumn{1}{|c|}{AUC} & \multicolumn{1}{|c|}{ACC} & \multicolumn{1}{|c|}{AUC} & \multicolumn{1}{|c|}{ACC} & \multicolumn{1}{|c|}{AUC} & \multicolumn{1}{|c}{ACC} \\
\midrule
PE                   & 0.788                   & 0.712                   & 0.790                    & 0.749                   & 0.848                   & 0.763                   \\
TE                   & 0.787                   & 0.712                   & 0.816                   & 0.766                   & 0.835                   & 0.758                   \\
RE                   & 0.755                   & 0.696                   & 0.686                   & 0.710                    & 0.835                   & 0.763                   \\
PE+TE                & 0.778                   & 0.705                   & 0.788                  & 0.746                   & 0.833                   & 0.754                   \\
PE+RE                & 0.759                   & 0.699                   & 0.676                   & 0.700                     & 0.832                   & 0.757                   \\
RE+TE                & 0.735                   & 0.692                   & 0.696                   & 0.705                   & 0.834                   & 0.757                   \\
PE+RE+TE             & 0.730                    & 0.684                   & 0.667                   & 0.693                   & 0.830                    & 0.756                   \\
RKT                  & \textbf{0.793}                   & \textbf{0.719}                   & \textbf{0.827  }                 & \textbf{0.774}                   & \textbf{0.860}                    & \textbf{0.770}     \\
\bottomrule
\end{tabular}
\label{ablation}
\end{table}
 
 To get deep insights on the RKT model, we investigate the contribution of various components involved in the model. 
 Therefore, we conduct some ablation experiments to
show how each part of our method affect final results. In Table ~\ref{ablation},
there are seven variations of RKT, each of which takes out one
or more opponents from the full model. Specifically:PE, TE, RE refer to RKT without position encoding, forget behavior modeling and exercise relation modeling, respectively. PE+TE, PE+RE, TE+RE refer to removal two components simultaneously, i.e. position encoding and
forget behavior modeling, position encoding and exercise relation modeling, and exercise relation modeling and forget behavior modeling,
respectively. And finally, PE+RE+TE refers to RKT that does not model the position encoding, forget behavior modeling and exercise relation modeling for interaction representation.
The result in Table ~\ref{ablation} indeed shows many interesting conclusions.\par
First, the more information a model encodes, the better the performance, which agrees with the intuition. Second for all datasets removing exercise relation modeling causes the most drastic drop in performance. This validates our argument that explicitly learning exercise relations is important for improving the performance of KT model. Thirdly, incorporating the forget behavior model in RKT which of students causes more improvement in ASSIST2012 and Junyi datasets than POJ. We hypothesize that this can be attributed to the fact that the concepts involved in solving POJ exercises are less diverse than those involved in high school maths course (Junyi and ASSIST2012 dataset). As a result in majority cases the reason of wrong answer on POJ  is the confusion in the students, rather than their forgetting behavior. 
\begin{table}[t]
\caption{Comparison of four exercise relation matrix computation methods.  }
\begin{tabular}{c|r|r|r|r|r|r}
\toprule
\multicolumn{1}{l}{} & \multicolumn{2}{|c|}{ASSIST2012}                    & \multicolumn{2}{|c|}{POJ}                           & \multicolumn{2}{|c}{Junyi}      \\
\midrule
\multicolumn{1}{l|}{} & \multicolumn{1}{c|}{AUC} & \multicolumn{1}{c|}{ACC} & \multicolumn{1}{c|}{AUC} & \multicolumn{1}{c|}{ACC} & \multicolumn{1}{c|}{AUC} & \multicolumn{1}{c}{ACC} \\
\midrule
Method (1)             & 0.755                   & 0.700                   & -                       & -                       & 0.764                   & 0.710                   \\
Method (2)             & 0.782                   & 0.708                   & 0.755                   & 0.733            &    0.836                   & 0.759                   \\
Method (3)             & 0.785                   & 0.709                   & 0.763                   & 0.737                   &   0.844                      &  0.762                      \\
Method (4)             & \textbf{0.793}                   & \textbf{0.719}                   & \textbf{0.827}                   & \textbf{0.774}                   & \textbf{0.860 }                  & \textbf{0.770}  \\
\bottomrule
\end{tabular}

\label{relation}
\end{table}

   \begin{figure*}[!t]
    {
       \includegraphics[width=0.95\textwidth]{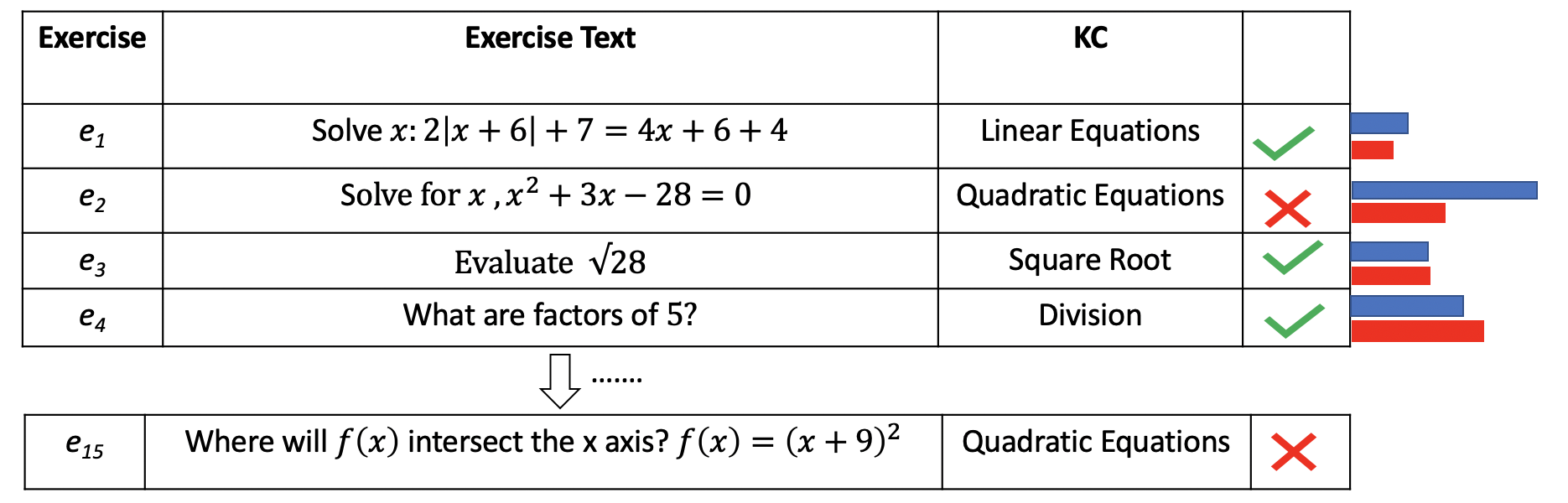}
     }
     \caption{Attention visualization in RKT model of an example student from Junyi. We predict her performance on $e_{15}$ based on her
past $15$ interaction (we only show the first $4$ interactions for better illustration). Right bars show the attention weights of two
RKT (blue) and SAKT (red) }
  \vspace{-2.8em}
     \label{vis}
   \end{figure*}

\begin{figure*}[!ht]
\label{heatmap}
     \subfloat[ ASSIST2012 - SAKT \label{subfloat-1}]{%
      \includegraphics[width=0.23\textwidth]{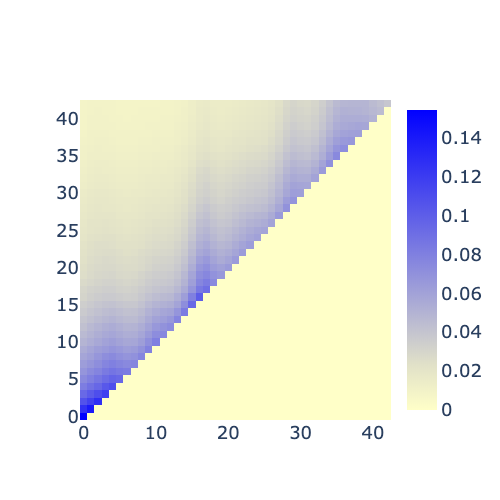}
    }
     \hfill
     \subfloat[ ASSIST2012 \label{subfloat-2}]{%
      \includegraphics[width=0.23\textwidth]{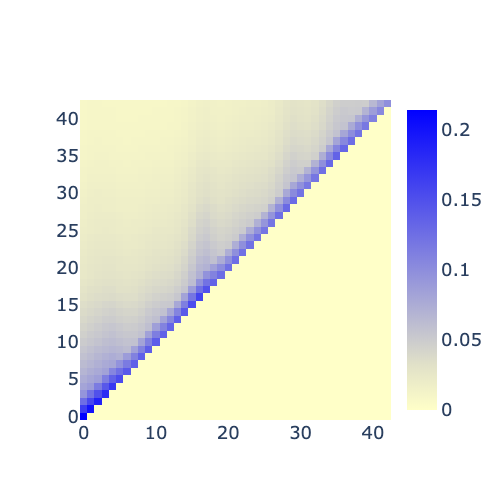}
     }
     \hfill
      \subfloat[ POJ \label{subfloat-3}]{%
      \includegraphics[width=0.23\textwidth]{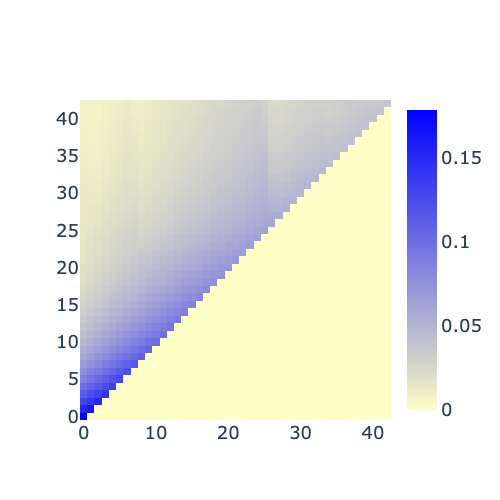}
     }
     \hfill
      \subfloat[Junyi\label{subfloat-3}]{%
      \includegraphics[width=0.23\textwidth]{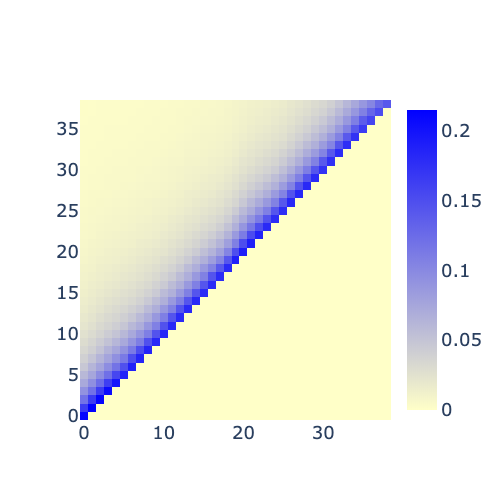}
     }s
     \caption{Visualization of attention weights on different datasets. Each subfloat depicts the average attention weights of different sequences of the corresponding datasets.  }
     \label{heatmap}
  \end{figure*}




\subsubsection{Effect of Exercise Relation matrix computation}
To explore the impact of exercise relation matrix computation, we consider the variants of RKT that uses different settings. We explore the following methods for computing exercise relation matrix:
\begin{enumerate}
\item Previous work such as ~\cite{lan2014sparse, 10.1145/3379507}, considered that two exercises are related if they belong to the same KC. We also employ this technique and build an exercise relation matrix with boolean values such that $\textbf{A}_{i,j}=1$ if $i$ and $j$ belong to the same KC otherwise $0$.
\item Use only the textual content of two  exercises to estimate the relation between them. We compute the relation between two exercises with Equation (3) only. 
    \item Use the student performance data to compute the relation between two exercises. Only Equation (2) is employed to compute the relation between two exercises.
    \item Use both textual content and student performance data to compute the similarity between two exercises. We compute the relation coefficients using Equation (4).
\end{enumerate}
We do not have information about the exercise-to-KC mapping for POJ data and hence can not apply method (1) for POJ.
Specifically Table ~\ref{relation} summarizes the experimental results. The findings are: \\
Firstly, Method (1) performs the worst among all the four methods. This can be attributed to the fact that linking exercises only based KCs ignores the fact there exists relation among exercises which do not belong to the same KC. Method (3) also shows performance gain over method (2) as student performance data is a good indicator of how relations between exercises are perceived by the students. Even if textual content of two exercises are not similar the association of knowledge involved in solving the two exercises could be high. 
Finally, method (4) that leverages both student performance data and exercise textual content data outperforms the other methods.

\vspace{-3.5mm}
 \subsection{Attention weights visualization (RQ3)}
 Benefiting from a purely attention mechanism, RKT and SAKT models are highly interpretable for explaining the prediction result.  To this end, we compared the attention weights obtained from both these models. We selected one student from Junyi dataset and obtain the attention weights corresponding to the past interactions for predicting her performance at exercise $e_{15}$.  Figure ~\ref{vis} shows the weights assigned by both SAKT and RKT. We see that compared to SAKT, RKT places more weights on $e_2$ which belongs to same KC as $e_{15}$ and have stronger relation. Since the student gave wrong answer to $e_2$, she has not yet mastered ``Quadratic Equations". As a result, RKT predicts that the student will not be able to answer $e_{15}$.
Thus, it is beneficial to consider relations between exercises for KT.\par
We also performed experiment to visualize the attention weights assigned by RKT on different datasets. Recall that at time step $t_i$, the relation-aware self-attention layer in our model revise the attention weights on the previous interactions depending on the time elapsed since the interaction  and the relations between the exercises involved. To this end, we examine all sequences and seek to reveal meaningful patterns by showing the average attention weights on the previous interactions.\par
 Figure ~\ref{heatmap} shows  the heatmap of attention weight matrix where $(i,j)$th element represents the attention weight on $j$th element when predicting performance at $i$th interaction. Note that when we calculate the average weight,
the denominator is the number of valid weights, so as to avoid
the influence of padding for short sequences.
We consider a few comparisons among the heatmaps:
\vspace{-3mm}
\begin{itemize}
\item (b), (c), (d): The heatmap representing the attention weights pertaining to different datasets reveals that recent interactions are given the higher weights compared to other interaction. It can be attributed to the forget behavior of learning process such that only the recent interactions can inform the student knowledge state.
 \item (b) vs. (c): This comparison shows the weights assigned by RKT on two different types of dataset. In ASSIST2012 dataset, the exercises are sequenced for \textit{skill-building}, i.e., they are organized so that a student can master one skill first and then learn the next skill. As a result in ASSIST2012 the exercises adjacent to each other are related. While, in POJ dataset, student chooses exercises based on their needs. As a result, the heatmap corresponding to ASSIST2012 dataset has attention weights concentrated towards the diagonal elements, while for POJ the attention weights are spread across the interactions.
\item (a) vs. (b): This comparison shows the effect of relation information for revising the attention weights. Without relation information the attention weights are more distributed over previous interaction, while the relation information concentrates the attention weights closer to diagonal as adjacent interactions in ASSIST2012 have higher relations.

\end{itemize}




\par
\section{Conclusion and Future Work}
In this work, we proposed a Relation-aware Self-attention mechanism for KT task, RKT. It models a student's interaction history and predicts her performance on the next exercise by considering contextual information obtained from its relation with the past exercises and the forget behavior of the student. The relation between exercises is computed using the student performance data and the textual content of exercises. The forget behavior is modeled using a time decaying kernel function. The contextual information is then incorporated in a self-attention layer which we call relation-aware self-attention. Extensive experimentation on real-world datasets shows that our model can outperform the state-of-the-art methods. Owing to the purely self-attention mechanism RKT is interpretable. \par
As part of future work, we plan to model the relation between exercises instead of computing them from the data. This can help in predicting the relation of a new exercise. Besides, we can learn a representation of student knowledge as an embedding and use this embedding to track student proficiency at various KCs. \par
\bibliographystyle{SIGCHI-Reference-Format}
\bibliography{ref}
\end{document}